\documentclass[10pt,twocolumn,letterpaper]{article}

\usepackage{cvpr}
\usepackage{times}
\usepackage{epsfig}
\usepackage{graphicx}
\usepackage{amsmath}
\usepackage{amssymb,gensymb}
\usepackage{nicefrac} 
\usepackage{tabulary,multirow,overpic}
\usepackage[dvipsnames]{xcolor}
\usepackage{array,booktabs}
\usepackage[caption=false]{subfig}
\usepackage{xspace}
\usepackage{wrapfig}
\usepackage{sidecap}

\newcommand\fyxsout{\bgroup\markoverwith{\textcolor{red}{\rule[0.5ex]{2pt}{1.2pt}}}\ULon}
\definecolor{mygray}{gray}{0.6}
\def\x{$\times$}
\newcommand{\figref}[1]{Fig.~\ref{#1}}
\newcommand{\tblref}[1]{Table~\ref{#1}}
\newcommand{\sref}[1]{Sec.~\ref{#1}}

\newcolumntype{x}[1]{>{\centering\arraybackslash}p{#1pt}}

\newlength\savewidth\newcommand\shline{\noalign{\global\savewidth\arrayrulewidth
  \global\arrayrulewidth 1pt}\hline\noalign{\global\arrayrulewidth\savewidth}}
\newcommand{\tablestyle}[2]{\setlength{\tabcolsep}{#1}\renewcommand{\arraystretch}{#2}\centering\footnotesize}
\makeatletter\renewcommand\paragraph{\@startsection{paragraph}{4}{\z@}
  {.5em \@plus1ex \@minus.2ex}{-.5em}{\normalfont\normalsize\bfseries}}\makeatother
\definecolor{demphcolor}{RGB}{100,100,100}
\newcommand{\demph}[1]{\textcolor{demphcolor}{#1}}
\newcommand{\app}{\raise.17ex\hbox{$\scriptstyle\sim$}}

\definecolor{citecolor}{RGB}{34,139,34}
\usepackage[pagebackref=true,breaklinks=true,letterpaper=true,colorlinks,
citecolor=citecolor,bookmarks=false]{hyperref}

\cvprfinalcopy 

\def\cvprPaperID{}

\begin{document}

\title{\vspace{-.5em} Audiovisual SlowFast Networks for Video Recognition \\ \vspace{-.5em}}

\author{
	\quad Fanyi Xiao\textsuperscript{1,2}\footnotemark  \quad
	Yong Jae Lee\textsuperscript{1} \quad
	Kristen Grauman\textsuperscript{2} \quad
	Jitendra Malik\textsuperscript{2} \quad
	Christoph Feichtenhofer\textsuperscript{2} \quad \vspace{.8em}\\
	\textsuperscript{1}University of California, Davis   \qquad\qquad 
	\textsuperscript{2}Facebook AI Research (FAIR)
}

\maketitle

\renewcommand*{\thefootnote}{\fnsymbol{footnote}}
\setcounter{footnote}{1}
\footnotetext{Work done during an internship at Facebook AI Research.}
\renewcommand*{\thefootnote}{\arabic{footnote}}
\setcounter{footnote}{0}

\definecolor{fastcolor}{RGB}{121,178,128}
\definecolor{slowcolor}{RGB}{165,170,243}
\newcommand{\fastcolor}[1]{\textcolor{fastcolor}{\textbf{#1}}}
\newcommand{\fastcolorC}[1]{\textcolor{orange}{#1}}
\newcommand{\slowcolor}[1]{\textcolor{slowcolor}{#1}}

\newcommand{\slow}{\slowcolor{Slow }}
\newcommand{\fast}{\fastcolor{Fast }}

\definecolor{predictioncolor}{RGB}{0,255,0}
\definecolor{labelcolor}{RGB}{255,0,0}
\newcommand{\predictioncolor}[1]{\textcolor{predictioncolor}{#1}}
\newcommand{\labelcolor}[1]{\textcolor{labelcolor}{#1}}

\newcommand{\pred}{\predictioncolor{\textbf{Predictions}: }}
\newcommand{\gt}{\labelcolor{\textbf{Labels}: }}

\begin{abstract}
\vspace{-5pt}
We present Audiovisual SlowFast Networks, an architecture for integrated audiovisual perception. AVSlowFast has Slow and Fast visual pathways that are deeply integrated with a Faster Audio pathway to model vision and sound in a unified representation. We fuse audio and visual features at multiple layers, enabling audio to contribute to the formation of hierarchical audiovisual concepts. To overcome training difficulties that arise from different learning dynamics for audio and visual modalities, we introduce DropPathway, which randomly drops the Audio pathway during training as an effective regularization technique. Inspired by prior studies in neuroscience, we perform hierarchical audiovisual synchronization to learn joint audiovisual features. We report state-of-the-art results on six video action classification and detection datasets, perform detailed ablation studies, and show the generalization of AVSlowFast to learn self-supervised audiovisual features. Code will be made available at: \url{https://github.com/facebookresearch/SlowFast}. 
\vspace{-5pt}

\end{abstract}

\vspace{-0.1in}
\section{Introduction}\label{sec:intro}

Joint audiovisual learning is core to human perception. However, most contemporary models for video analysis  exploit only the visual signal and ignore the audio signal. For many video understanding tasks,  
{audio} could be very helpful.    
{Audio has the potential to influence action recognition not only in  obvious cases where sound dominates, like ``playing saxophone", but also visually subtle cases where the action itself is is difficult to see in the video frames, like ``whistling", or closely related actions where sound helps disambiguate, like ``closing" vs. ``slamming" the door.}

\begin{figure}[t!]
    \centering
    \vspace{5pt}
    \hspace*{-6pt}
    \includegraphics[width=0.49\textwidth]{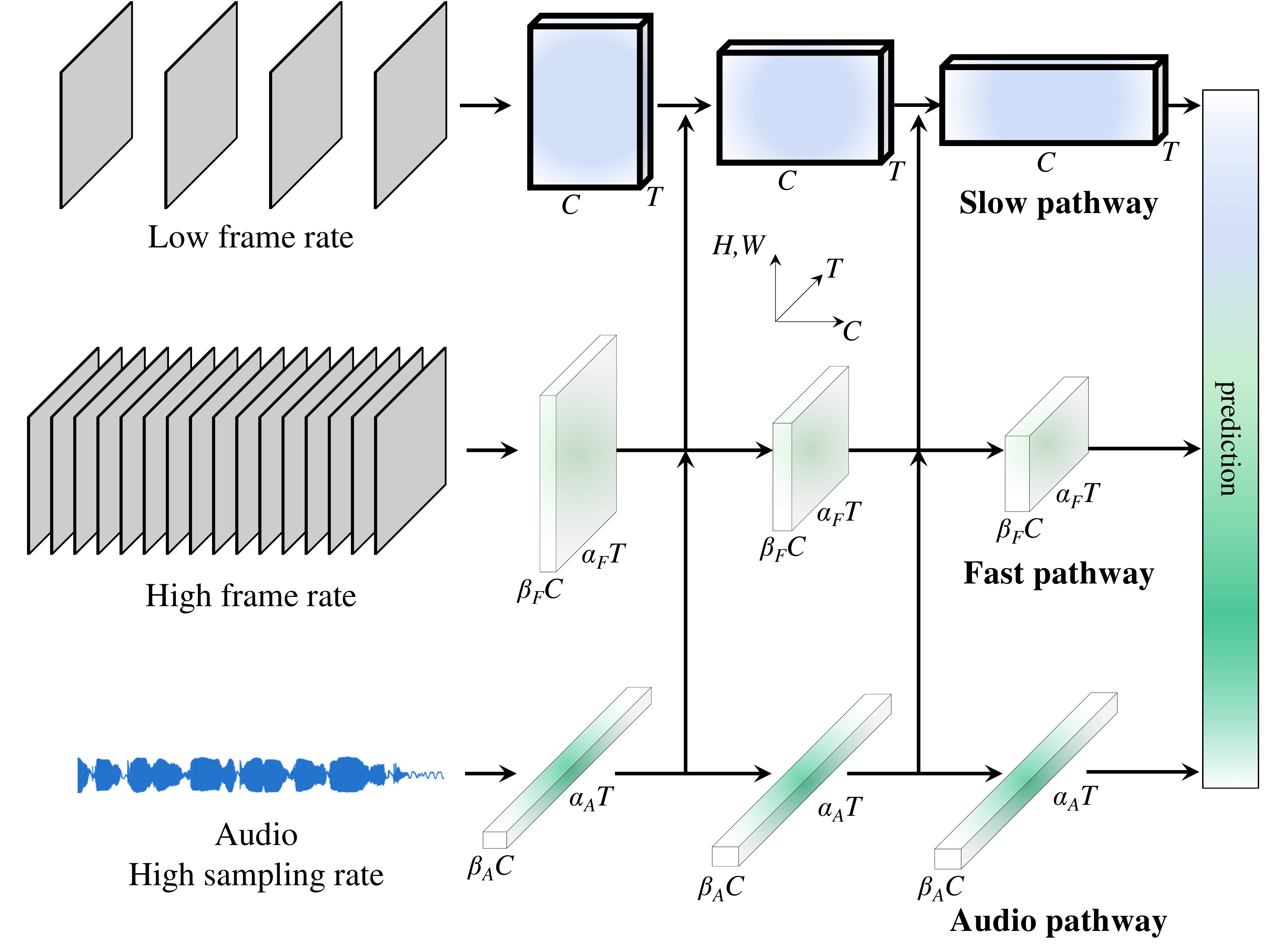}
    \caption{\small \textbf{Audiovisual SlowFast Networks}
		 have Slow and Fast visual pathways that are deeply integrated with a Faster Audio pathway  to model vision and sound in a unified representation. 
	}
    \vspace{-15pt}
    \label{fig:teaser}
\end{figure}

This line of thinking is supported by perceptual and neuroscience studies suggesting interesting ways in which visual and audio signals are combined in the brain.  A classic example is the McGurk effect~\cite{mcgurk-nature1976}\footnote{\url{https://www.youtube.com/watch?v=G-lN8vWm3m0}} -- when one is listening to an audio clip (e.g., sounding ``ba-ba''), alongside watching a video of fabricated lip movements (indicating ``va-va''), the sound one perceives \textit{changes} (in this case from ``ba-ba'' to ``va-va''). 

This effect demonstrates that there is tight entanglement between audio and visual signals (known as the \emph{\mbox{multisensory} integration process})~\cite{massaro-acoustic2000,keysers-audiovisual2003,stein-multi2012,sotofaraco-book2019}. Importantly, research has suggested this fusion between audio and visual signals happens at a fairly early stage~\cite{schwartz-audio2002,omata-fusion2007}.

Given its high potential in facilitating video understanding, researchers have attempted to utilize audio in videos~\cite{kazakos-epicfusion2019,ghanem-activitynet2018,arandjelovic-iccv2017,aytar-nips2016,owens-eccv2018,owens-eccv2016,arandjelovic-eccv2018,senocak-cvpr2018,gao-eccv2018}. However, there are a few  challenges in making effective use of audio. First, audio does not always correspond to the visual frames (e.g., in a {``dunking basketball''} video, there can be class-unrelated background music playing). Conversely, audio does not always contain information that can help understand the video (e.g., {``shaking hands''} does not have a particular sound signature).
There are also challenges from a technical perspective.  Specifically, we identify the incompatibility of ``learning dynamics'' between the visual and audio pathways -- audio pathways generally train much faster than visual ones, which can lead to generalization issues during joint audiovisual training. 
Due in part to these various difficulties, a principled approach for audiovisual modeling is currently lacking.  Many previous methods adopt an ad-hoc scheme that consists of a separate audio network that is integrated with the visual pathway via ``late-fusion''~\cite{ghanem-activitynet2018,arandjelovic-iccv2017,owens-eccv2018}. 

The objective of this paper is to build an architecture for \textit{integrated audiovisual perception}.  We aim to go beyond previous work that performs ``late-fusion'' of independent audio and visual pathways, to instead learn \emph{hierarchies} of integrated audiovisual features, enabling unified audiovisual perception.
We propose a new architecture, \emph{\mbox{Audiovisual} \mbox{SlowFast}} Networks (\mbox{AVSlowFast}), to perform fusion at multiple levels (Fig.~\ref{fig:teaser}). AVSlowFast Networks build on 
{SlowFast}~\cite{slowfast}, a class of architectures that has two pathways, of which one (Slow) is designed to capture more static but semantic-rich information whereas the other (Fast) is tasked to capture motion. AVSlowFast hierarchically intertwines a Faster Audio pathway with the Slow and Fast pathways, as audio has higher sampling rate, that learns end-to-end from vision and sound. The Audio pathway can be lightweight ($<$20\% of computation), but requires  a careful design and training strategies to be useful in practice.

We evaluate our approach on standard datasets in the human action recognition community and  find consistent improvement for integrating audio. The improvement in accuracy varies for datasets and classes but comes with a relatively small increase in computational cost. For example, on the leading dataset for egocentric video, EPIC-kitchens~\cite{damen-epic2018}, audio boosts by +2.9/+4.3/+2.3 the top-1 accuracy for verb/noun/action recognition at 20\%  of overall compute, on Kinetics~\cite{kay-kinetics2017} action classification by +1.4 top-1 accuracy at 11\% of compute, and on AVA~\cite{gu-ava2018}  action detection by +1.2 mAP at only 2\%  of the overall compute. 

Our key contributions are:

(i) We present AVSlowFast, {which} 
fuses audio and visual information \emph{at multiple levels} in the network hierarchy (\ie, hierarchical fusion) so that audio can contribute to the formation of visual concepts at different levels of abstraction.  In contrast to late-fusion, this enables the audio signal to participate in the process of forming visual features. 

(ii) To overcome the incompatibility of learning dynamics between the visual and audio pathways, we propose \emph{DropPathway}, {which}
randomly drops the Audio pathway during training as a simple and effective regularization technique to tune the pace of the learning process. This enables us to train our joint audiovisual model with hierarchical fusion connections across modalities. 

(iii) Inspired by the \mbox{multisensory} integration process mentioned above and prior work in neuroscience~\cite{keysers-audiovisual2003}, which suggests that there exist \emph{audiovisual mirror neurons} in monkey brains that respond to ``any evidence of the action, be it auditory or visual'', we propose to perform audio visual synchronization (AVS)~\cite{owens-eccv2018,korbar-nips2018,arandjelovic-iccv2017,aytar-nips2016} at multiple layers to learn features that generalize across modalities.

(iv) We conduct extensive experiments on six video recognition datasets for human action classification and detection. We report state-of-the-art results and provide ablation studies to understand {the} trade-offs of various design choices.  In addition to evaluating the performance of AVSlowFast for established supervised video classification and detection tasks, we validate the generalization of the audiovisual representation to self-supervised learning, revealing that strong video features can be learned with AVSlowFast using standard pretraining objectives.

\section{Related Work}\label{sec:relatedwork}\vspace{-5pt}

\paragraph{Video recognition.} 
Significant progress has been made in video recognition in recent years. Some notable directions are two-stream networks in which one stream processes RGB frames and the other processes optical flow~\cite{simonyan-nips2014,feichtenhofer-nips2016,wang-eccv2016}, 3D ConvNets as an extension of 2D networks to the spatiotemporal domain~\cite{tran-iccv2015,qiu-iccv2017,xie-eccv2018}, and recent SlowFast Networks that have two pathways to process videos at different temporal frequencies~\cite{slowfast}. Despite all these efforts on harnessing temporal information in videos, research is relatively lacking when it comes to another important information source -- audio in video. 

\paragraph{Audiovisual activity recognition.}

Joint modeling of audio and visual signals has been largely conducted in a ``late-fusion'' manner in video recognition literature~\cite{kazakos-epicfusion2019,long-cvpr2018,ghanem-activitynet2018}. For example, all the entries that utilize audio in the 2018 ActivityNet challenge report~\cite{ghanem-activitynet2018} have adopted this paradigm -- meaning that there are networks processing visual and audio inputs separately, and then they either concatenate the output features or average the final class scores across modalities.     Recently, an interesting audiovisual fusion approach has been proposed~\cite{kazakos-epicfusion2019} using flexible binding windows when fusing audio and visual features. 
With three similar network streams, this approach fuses audio features with the features from RGB and optical flow at the final stage before classification. 
In contrast, AVSlowFast is building a hierarchically integrated audiovisual representation.

\paragraph{Multi-modal learning.}
Researchers have long been interested in developing models that can learn from multiple modalities (e.g., audio, vision, language). Beyond audio and visual modalities, extensive research has been conducted in other instantiations of multi-modal learning, {including vision and language~\cite{fukui-multimodal2016,arevalo-multimodal2017,xiao-cvpr2017}, vision and locomotion~\cite{zhu-icra2017,gandhi-iros2017}, and learning from physiological data~\cite{laurent-multimodal2013}.}

\paragraph{Other audiovisual tasks.}
Audio has also been extensively utilized outside of video recognition, \eg for learning audiovisual representations in a self-supervised manner~\cite{arandjelovic-iccv2017,aytar-nips2016,owens-eccv2018,owens-eccv2016,korbar-nips2018,hu-cvpr2019} by exploiting audio-visual correspondence.
Other audiovisual tasks include audio-visual speech recognition~\cite{potamianos-avsr2003,ngiam-multimodal2011}, lip reading \cite{chung2017lip}, biometric matching~\cite{nagrani-cvpr2018}, sound-source localization~\cite{casanovas-multimedia2010,arandjelovic-eccv2018,senocak-cvpr2018,zhao-eccv2018,hu-cvpr2019,ephrat-arxiv2018}, audio-visual source separation~\cite{owens-eccv2018,gao-eccv2018}, and audiovisual question answering~\cite{alamri-aaaiw2018}.

\section{Audiovisual SlowFast Networks}
Inspired by research in neuroscience~\cite{bernstein-neuro2013}, which suggests that audio and visual signals fuse at multiple levels, we propose to fuse audio and visual features at multiple stages, from intermediate-level features to high-level semantic concepts. This way, audio can participate in the formation of visual concepts at different levels. 
AVSlowFast Networks are conceptually simple: SlowFast has Slow and Fast pathways to process visual input (\S\ref{sec:slowfastpathways}), and AVSlowFast extends this with an Audio pathway (\S\ref{sec:audiopathway}).  

\subsection{SlowFast pathways}\label{sec:slowfastpathways}
We begin by briefly reviewing the SlowFast architecture~\cite{slowfast}. The Slow pathway (Fig.~\ref{fig:teaser}, top row) is a convolutional network that processes videos with a large temporal stride (\ie, it samples one frame out of $\tau$ frames). The primary goal of the Slow pathway is to produce features that capture semantic contents of the video, which has a low {refresh} rate (semantics do not change all of a sudden). The Fast pathway (Fig.~\ref{fig:teaser}, middle row) is another convolutional model with three key properties.  First, it has an $\alpha_F$ times higher frame rate (\ie, with temporal stride $\tau / \alpha_F$, $\alpha_F > 1$) so that it can capture fast motion information. Second, it preserves fine temporal resolution by avoiding any temporal downsampling. Third, it has a lower channel capacity ($\beta_F$ times the Slow pathway channels, where $\beta_F<1$) as it is demonstrated to be a desired trade-off~\cite{slowfast}. We refer readers to~\cite{slowfast} for more details.   

\subsection{Audio pathway}\label{sec:audiopathway}

A key property of the Audio pathway is that it has an even finer temporal structure than the Slow and Fast pathways (with waveform sampling rate on the order of kHz). As standard processing, we take a log-mel-spectrogram (2-D representation in time and frequency of audio) as input and set the temporal stride to $\tau / \alpha_A$ frames, where $\alpha_A$ can be much larger than $\alpha_F$ (\eg, 32 \vs 8).  In a sense, it serves as a ``Faster'' pathway with respect to Slow and Fast pathways. Another notable property of the Audio pathway is its low computation cost, as audio signals, due to their lower-dimensional nature, are cheaper to process than visual signals. 
To control this, we set the channels of the Audio pathway to $\beta_A$ \x~Slow pathway channels. By default, we set $\beta_A$ to 1$/$2. Depending on the specific instantiation, the Audio pathway typically only requires 10\% to 20\% of the overall computation of AVSlowFast. 

\subsection{Lateral connections} 

In addition to the lateral connections between the Slow and Fast pathways in~\cite{slowfast}, we add lateral connections between the Audio, Slow, and Fast pathways to fuse audio and visual features.  Following~\cite{slowfast}, lateral connections are added after ResNet ``stages'' (\eg, pool$_1$, res$_2$, res$_3$, res$_4$ and pool$_5$). However, unlike~\cite{slowfast}, which has lateral connections after each stage, we found that it is most beneficial to have lateral connections between audio and visual features starting from intermediate levels (we ablate this in Sec.~\ref{sec:ablation}). This is conceptually intuitive as very low-level visual features such as edges and corners might not have a particular sound signature. 
 Next, we discuss several concrete AVSlowFast instantiations.

\subsection{Instantiations}\label{sec:instantiations}

AVSlowFast Networks define a generic class of models that follow the same design principles. In this section, we exemplify a specific instantiation in Table~\ref{table:arch}. 
We denote spatiotemporal size by $T$\x $S^2$ for Slow/Fast pathways and $F$\x$T$ for the Audio pathway, where $T$ is the temporal length, $S$ is the height and width of a square spatial crop, and $F$ is {the number of} frequency bins for audio.

\newcommand{\blocks}[3]{\multirow{3}{*}{\(\left[\begin{array}{c}\text{1$\times$1$^\text{2}$, #2}\\[-.1em] \text{1$\times$3$^\text{2}$, #2}\\[-.1em] \text{1$\times$1$^\text{2}$, #1}\end{array}\right]\)$\times$#3}
}
\newcommand{\blockt}[3]{\multirow{3}{*}{\(\left[\begin{array}{c}\text{{3$\times$1$^\text{2}$}, #2}\\[-.1em] \text{1$\times$3$^\text{2}$, #2}\\[-.1em] \text{1$\times$1$^\text{2}$, #1}\end{array}\right]\)$\times$#3}
}

\newcommand{\blocka}[3]{\multirow{3}{*}{\(\left[\begin{array}{c}\text{1$\times$1, #2}\\[-.1em] \text{3$\times$3, #2}\\[-.1em] \text{1$\times$1, #1}\end{array}\right]\)$\times$#3}
}

\newcommand{\blocktf}[3]{\multirow{3}{*}{\(\left[\begin{array}{c}\text{1$\times$1, #2}\\[-.1em] \text{[3$\times$1, 1$\times$3], #2}\\[-.1em] \text{1$\times$1, #1}\end{array}\right]\)$\times$#3}
}

\newcommand{\outsizes}[7]{\multirow{#7}{*}{\(\begin{array}{c} \text{\emph{Slow}}: \text{#1$\times$#2$^\text{2}$}\\[-.1em] \text{\emph{Fast}}: \text{#4$\times$#5$^\text{2}$}\end{array}\)}
}

\begin{table}[t]
	\scriptsize
	\centering
	\resizebox{\columnwidth}{!}{
		\tablestyle{1pt}{1.08}
		\vspace{-10pt}
		\begin{tabular}{c|c|c|c}
			stage & \emph{Slow} pathway &  \emph{Fast} pathway & \emph{Audio} pathway \\
			\shline
			\multirow{1}{*}{raw clip} & 3\x64\x224$^2$ & 3\x64\x224$^2$ & 80\x128 (\emph{freq.}{\x}\emph{time}) \\
			\hline
			\multirow{2}{*}{data layer} & \multirow{2}{*}{stride 16, 1$^\text{2}$} & \multirow{2}{*}{stride 2, 1$^\text{2}$} &  \multirow{2}{*}{-}   \\
			&  &  \\
			\hline
			\multirow{2}{*}{conv$_1$} & \multicolumn{1}{c|}{1\x7$^\text{2}$, {64}} & \multicolumn{1}{c|}{{5\x7$^\text{2}$}, 8} &  \multicolumn{1}{c}{[9\x1, 1\x9], 32}   \\
			& stride 1, 2$^\text{2}$ & stride 1, 2$^\text{2}$ & stride 1, 1 \\
			\hline
			\multirow{2}{*}{pool$_1$}  & \multicolumn{1}{c|}{1\x3$^\text{2}$ max} & \multicolumn{1}{c|}{1\x3$^\text{2}$ max} & \multirow{2}{*}{-} \\
			& stride 1, 2$^\text{2}$ & stride 1, 2$^\text{2}$ & \\
			\hline
			\multirow{3}{*}{res$_2$} & \blocks{{256}}{{64}}{3} & \blockt{{32}}{{8}}{3} & \blocktf{{128}}{{32}}{3}  \\
			&  & \\
			&  & \\
			\hline
			\multirow{3}{*}{res$_3$} & \blocks{{512}}{{128}}{4} &  \blockt{{64}}{{16}}{4}  & \blocktf{{256}}{{64}}{4}  \\
			&  & \\
			&  & \\
			\hline
			\multirow{3}{*}{res$_4$} & \blockt{{1024}}{{256}}{6} & \blockt{{128}}{{32}}{6} &  \blocka{{512}}{{128}}{6}  \\
			&  & \\
			&  & \\
			\hline
			\multirow{3}{*}{res$_5$} & \blockt{{2048}}{{512}}{3} & \blockt{{256}}{{64}}{3} &   \blocka{{1024}}{{256}}{3} \\
			&  & \\
			&  & \\
			\hline
			\multicolumn{4}{c}{global average pool, concat, fc} \\
		\end{tabular}}
	\vspace{.1em}
	\caption{\textbf{An instantiation of the AVSlowFast network}. 
		For Slow \& Fast pathways, the dimensions of kernels are denoted by $\{$$T$\x $S^2$, $C$$\}$ for temporal, spatial, and channel sizes. For the Audio pathway, kernels are denoted with $\{$$F${\x}$T$, $C$$\}$, where $F$ and $T$ are frequency and time.   
		Strides are denoted with $\{$temporal stride, spatial stride$^2$$\}$ and $\{$frequency stride, time stride$\}$ for SlowFast and Audio pathways, respectively.
		The speed ratios are $\alpha_F=$ 8, $\alpha_A=$ 32 and the channel ratios are $\beta_F=$1$/$8, $\beta_A=$1$/$2 and \mbox{$\tau$ $=$ 16}. The backbone is \mbox{ResNet-50}.
	}
	\label{table:arch}
\end{table}

\paragraph{SlowFast pathways.}
For Slow and Fast pathways, we follow the basic instantiation of SlowFast 4\x16, R50 model defined in~\cite{slowfast}. It has a Slow pathway that samples $T=4$ frames out from a 64-frame raw clip with a temporal stride $\tau=16$. There is no temporal downsampling in the Slow pathway, since input stride is large. Also, it only applies non-degenerate temporal convolutions (temporal stride $>1$) in res$_4$ and res$_5$ (see Table~\ref{table:arch}), as this is more effective. 

For the Fast pathway, it has a higher frame rate ($\alpha_F=8$) and a lower channel capacity ($\beta_F=1/8$), such that it can better capture motion while trading off model capacity. To preserve fine temporal resolution, the Fast pathway has non-degenerate temporal convolutions in every residual block. 
Spatial downsampling is performed with stride 2$^\text{2}$ convolution in the center (``bottleneck'') filter of the first residual block in each stage of both the Slow and Fast pathways. 

\paragraph{Audio pathway.}

The Audio pathway takes as input the log-mel-spectrogram representation, which is a 2-D representation with one axis being time and the other one denoting frequency bins. In Table~\ref{table:arch}, we use 128 spectrogram frames (corresponding to 2 seconds of audio) with 80 bins.  

Similar to the Slow and Fast pathways, the Audio pathway is also based on a ResNet, but with specific design to better fit the audio inputs.  
First, we do not pool after the initial convolutional filter  (\ie there is no downsampling layer at stage pool$_1$) to preserve information along both temporal and frequency axis.
Downsampling in time-frequency space is performed by stride 2$^\text{2}$ convolution in the center (``bottleneck'') filter of the first residual block in each stage from res$_2$ to res$_5$.
  Second, we decompose the 3{\x}3 convolution filters in res$_2$ and res$_3$ into 1\x3 filters for frequency and 3\x1 filters for time. This increases accuracy slightly (by 0.2\% on Kinetics) but also reduces computation. Conceptually, it allows the network to treat time and frequency separately (as opposed to 3\x3 filters which imply both axes are uniform) in  early stages. 
While for spatial filters it is reasonable to perform filtering in $x$ and $y$ dimensions symmetrically, this might not be optimal for early filtering in time and frequency dimensions, as the statistics of spectrograms are different from natural images, which instead are approximately isotropic and shift-invariant \cite{Ruderman1994,Huang1999}. 

\paragraph{Lateral connections.}
There are many options for how to fuse audio features into the visual pathways. Here, we describe several instantiations and the motivation behind them. Note that this section discusses the lateral connections between the Audio and SlowFast pathways. For the fusion connection between the two visual pathways (Slow and Fast), we adopt the temporal strided convolution as it is demonstrated to be most effective in~\cite{slowfast}. 

\begin{figure}[t!]
	 \vspace{-12pt}
	\hspace*{-5pt}
    \centering
    \includegraphics[width=0.475\textwidth]{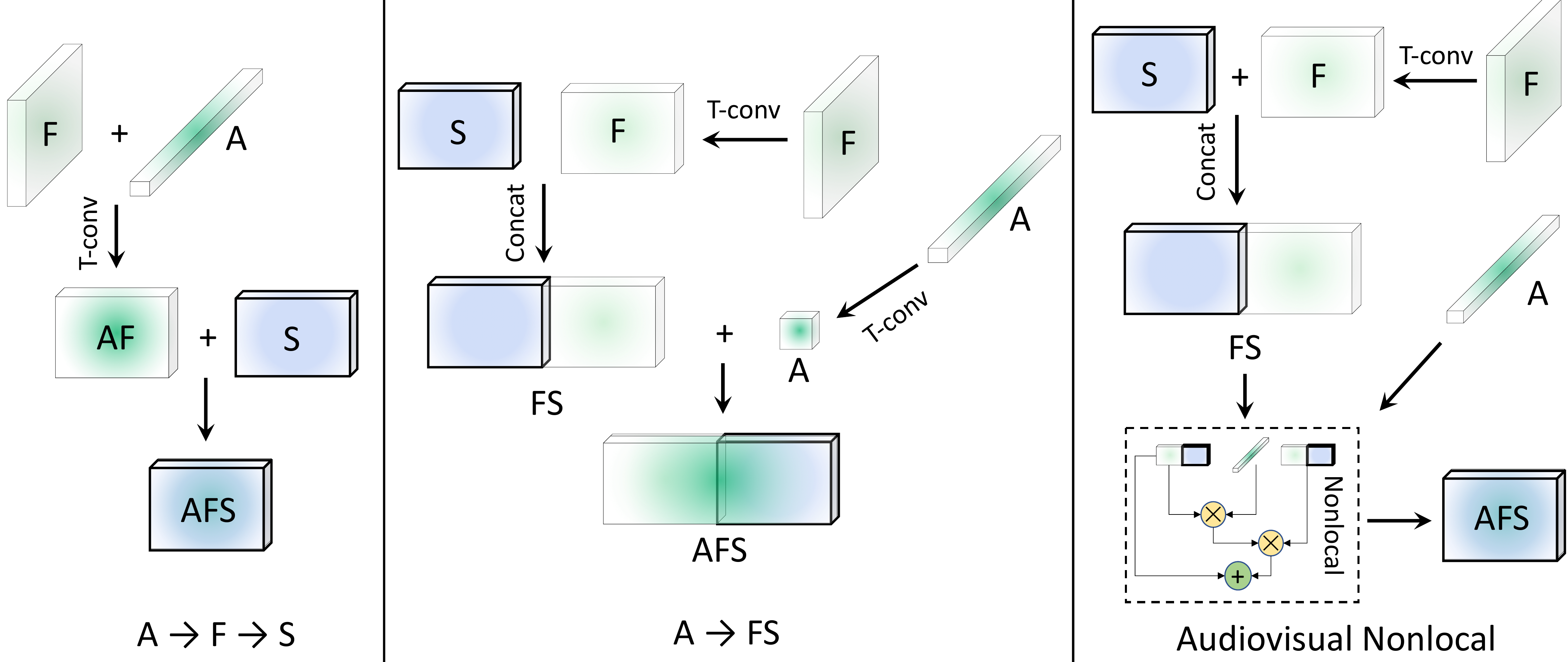}
    \caption{\small \textbf{Fusion connections for AVSlowFast}. \textit{Left}: A$\rightarrow$F$\rightarrow$S enforces strong temporal alignment between audio and RGB frames, as audio is fused into the Fast pathway with fine temporal resolution. \textit{Center}: A$\rightarrow$FS has higher tolerance on temporal misalignment as audio is fused into the temporally downsampled output of SlowFast fusion. \textit{Right}: Audiovisual Nonlocal fuses through a Nonlocal block \cite{wang-nonlocal2018}, such that audio features are used to select visual features that are deemed important by audio.\vspace{-5pt}  
    }
    \label{fig:fusion}
\end{figure}

\noindent  \textbf{(i) \emph{A$\rightarrow$F$\rightarrow$S}}:
In this approach (Fig.~\ref{fig:fusion} left), the Audio pathway (A) is first fused to the Fast pathway (F), and then fused to the Slow pathway (S). Specifically, audio features are subsampled to the temporal length of the Fast pathway and then fused into the Fast pathway with a \textit{sum} operation. After that, the resulting features are further subsampled by $\alpha_F$ (e.g., ~4\x~subsample) and fused with the Slow pathway (as is done in SlowFast). The key property of this approach is that it enforces \emph{strong temporal alignment} between audio and visual features, as audio features are fused into the Fast pathway which preserves fine temporal resolution. 
  
\noindent   \textbf{(ii) \emph{A$\rightarrow$FS}}: 
An alternative way is to fuse the Audio pathway into the output of the SlowFast fusion (Fig.~\ref{fig:fusion} center), which is coarser in temporal resolution. We adopt this design as our default choice as it imposes a less stringent requirement on temporal alignment between audio and visual features, which we found to be important in our experiments. Similar ideas of relaxing the alignment requirement are also explored in \cite{kazakos-epicfusion2019},  in the context of combining RGB, flow, and audio streams. 

\noindent \textbf{(iii) \emph{Audiovisual Nonlocal}}: 
One might also be interested in using audio as a \emph{modulating signal} to visual features. Specifically, instead of directly summing or concatenating audio features into the visual stream, one might expect audio to play a more subtle role of modulating the visual concepts, through attention mechanisms such as Non-Local (NL) blocks~\cite{wang-nonlocal2018}. One example would be audio serving as a probing signal indicating where the interesting event is happening in the video, both spatially and temporally, and then focusing the attention of visual pathways on those locations. To materialize this, we adapt NL blocks to take both audio and visual features as inputs (Fig.~\ref{fig:fusion} right). Audio features are then matched to different locations within visual features (along $H$, $W$ and $T$ axis), and the affinity is used to generate a new visual feature that combines information from locations deemed important by audio features. 

\vspace{-5pt}
\subsection{Joint audiovisual training} 

Unlike SlowFast, AVSlowFast trains with multiple modalities. As noted in Sec.~\ref{sec:intro}, this leads to challenging training dynamics (\ie, different training speed of audio and visual pathways). To tackle this, we propose two training strategies that enable joint training.  

\begin{figure}[t!]
    \centering
    \vspace{-10pt}
    \includegraphics[width=0.35\textwidth]{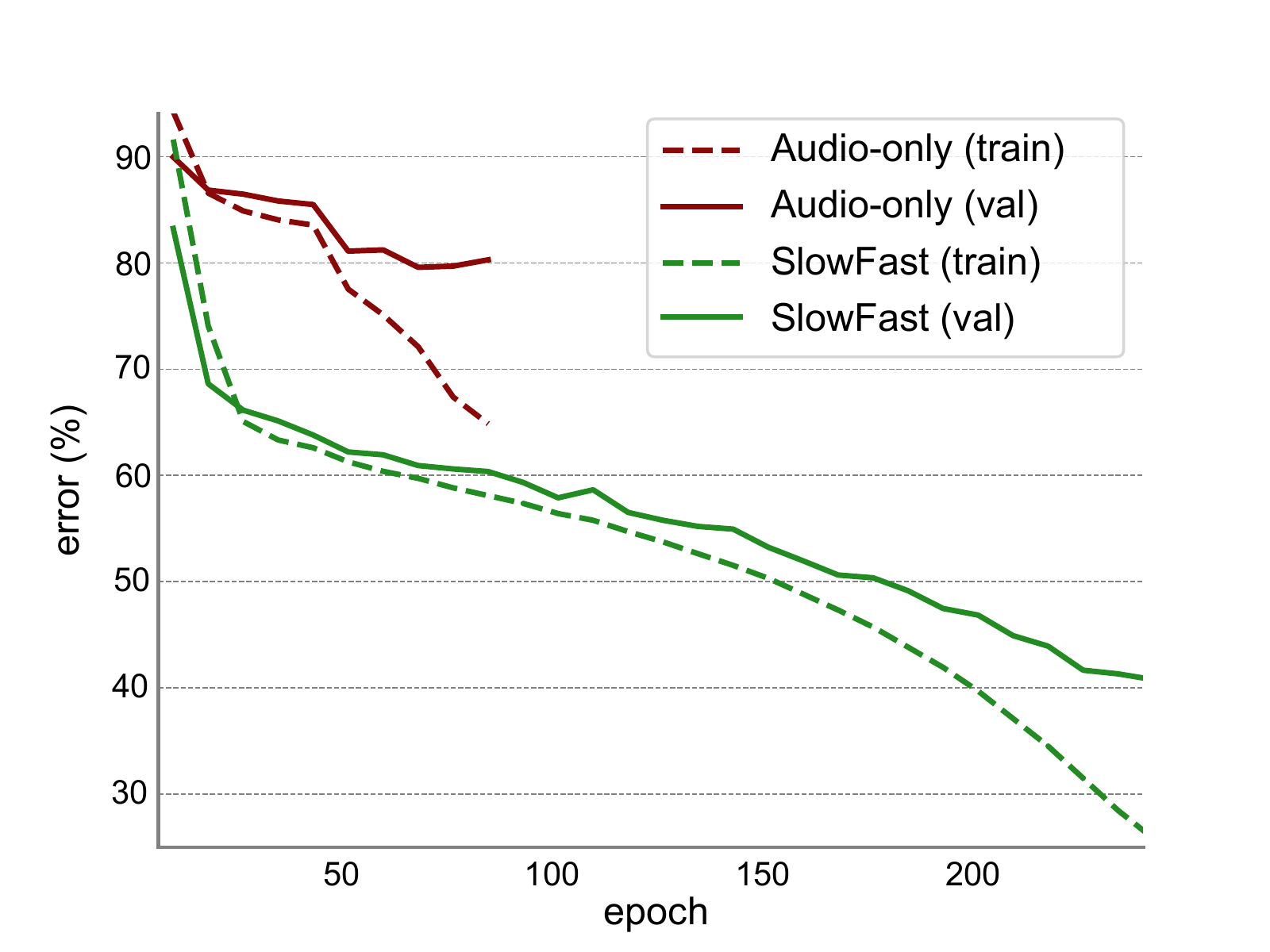}
    \caption{Training procedure on Kinetics for Audio-only (red) \vs SlowFast (green) networks. We show the top-1 training error (dash) and validation error (solid). The curves show single-crop \emph{errors}; the video \emph{accuracy} is 24.8\% \vs 75.6\%. The audio network converges after around 3\x~fewer iterations compared to the visual. }
    \label{fig:overfit}
    \vspace{-12pt}
\end{figure}

\paragraph{DropPathway.} 
We discuss a possible reason for why previous methods employ audio in a late fusion approach. By analyzing the model training dynamics we observe the following. Audio and visual pathways are very different in terms of their ``learning speed''. 

Taking the curves in Fig.~\ref{fig:overfit} as an example, the green curve is for training a visual-only SlowFast model, whereas the red curve is for training an Audio-only model. It shows that the Audio-only model requires fewer training iterations before it starts to overfit (at $\app$70 epochs, which is $\app1/3$ of the visual model's training epochs). 
One modality dominating multi-modal training has also been observed for lip-reading applications \cite{chung2017lip} and optical flow streams in action recognition  \cite{feichtenhofer-cvpr2016} {and video object segmentation~\cite{fusionseg}.} 

The discrepancy on learning pace leads to overfitting if we naively train both modalities jointly. To unlock the potential of joint training, we propose a simple strategy of {randomly dropping the Audio pathway during training} (\textit{DropPathway}).
Specifically, at each training iteration, we drop the Audio pathway altogether with probability ${P}_{d}$. This way, we \emph{slow down} the learning of the Audio pathway and make its learning dynamics more compatible with its visual counterpart. When dropping the audio pathway, we sum zero tensors with the visual pathways (we also explored feeding the running average of audio features, and found similar results, possibly due to BN). 

Our ablation studies in the next section will show the effect of DropPathway, showing that this simple strategy provides good generalization and is essential for jointly training AVSlowFast. Note that DropPathway is different {from simply} setting different learning rates for the audio/visual pathways in that it 1) ensures the Audio pathway has fewer parameter updates, 2) hinders the visual pathway to `shortcut' training by memorizing audio information, and 3) provides extra regularization as different audio clips are dropped in each epoch. 

\paragraph{Hierarchical audiovisual synchronization.}
As noted in Sec.~\ref{sec:relatedwork}, temporal synchronization (that comes for free) between audio and visual sources has been explored as a self-supervisory signal to learn feature representations~\cite{owens-eccv2016,aytar-nips2016,arandjelovic-iccv2017,chung-accv2016,owens-eccv2018,korbar-nips2018,han-dpc2019}. In this work, we use audiovisual synchronization to encourage the network to produce feature representations that are generalizable across modalities (inspired by the \emph{audiovisual mirror neurons} in primate vision~\cite{keysers-audiovisual2003}). Specifically, we add an auxiliary task to classify whether a pair of audio and visual frames are \emph{in-sync} or not~\cite{korbar-nips2018,owens-eccv2018} and adopt a curriculum schedule used in~\cite{korbar-nips2018} that starts with easy negatives (audio and visual frames come from different clips), and transition into a mix of easy and hard (audio and visual frames are from the same clip, but with a temporal shift) after 50\% of training epochs. 
In our experiments, we study the effect of audiovisual synchronization for both supervised and self-supervised audiovisual feature learning.

\section{Experiments: Action Classification} \label{sec:kinetics}

We evaluate our approach on six video recognition datasets using standard evaluation protocols. For the action classification experiments in this section we use EPIC-Kitchens~\cite{damen-epic2018}, Kinetics-400~\cite{kay-kinetics2017}, and Charades~\cite{sigurdsson-charades2016}. For action detection, we use the AVA dataset~\cite{gu-ava2018} covered in \sref{sec:ava}, and the AVSlowFast self-supervised representation is evaluated on UCF101~\cite{soomro-ucf2012} \& HMDB51~\cite{kuehne-hmdb2011} in \sref{sec:ssl}.
	
\noindent\textbf{Datasets.} 
The EPIC-Kitchens dataset~\cite{damen-epic2018} consists of daily activities captured in various kitchen environments with egocentric video and sound recordings. It has 39k segments in 432 videos. For each segment, the task is to predict a verb (\eg, ``turn-on''), a noun (\eg, ``switch''), and an action by combining the two (``turn on switch''). Performance is measured as top-1 and top-5 accuracy. We use the train/val split in~\cite{baradel-eccv2018}. Test results are obtained from the evaluation server.
	
Kinetics-400~\cite{kay-kinetics2017} (abbreviated as K400) is a large-scale video dataset of $\app$240k training videos and 20k validation videos in 400 action categories. Results on Kinetics are reported as top-1 and top-5 classification accuracy (\%). 

Charades~\cite{sigurdsson-charades2016} is a dataset of $\app$9.8k training videos and 1.8k validation videos in 157 classes. Each video has multiple labels of activities spanning $\app$30 seconds. Performance is measured in mean Average Precision (mAP).
	
\noindent\textbf{Audio pathway.} 
Following previous work~\cite{arandjelovic-iccv2017,arandjelovic-eccv2018,korbar-nips2018,korbar-scsampler2019}, we extract log-mel-spectrograms from the raw audio waveform to serve as the input to Audio pathway. Specifically, we sample audio data with 16 kHz sampling rate, then compute a spectrogram with window size of 32ms and step size of 16ms. 
The length of the audio input is exactly matched to the duration spanned by the RGB frames. For example, under 30 FPS, for AVSlowFast with $T$\x$\tau$  $=$ 8\x8 frames (2 secs) input, we sample 128 frames (2 secs) in log-mel.

\noindent\textbf{Training.} We train our AVSlowFast models on Kinetics from scratch without pre-training. We use synchronous SGD  and follow the training recipe (learning rate, weight decay, warm-up) used in~\cite{slowfast}. Given a training video, we randomly sample $T$ frames with stride $\tau$ and extract the corresponding log-mel-spectrogram. We randomly crop 224\x224 pixels from a video, randomly flip horizontally, and resize it to a shorter side sampled in [256, 320]. 

\noindent\textbf{Inference.}
Following previous work~\cite{wang-nonlocal2018,slowfast}, we uniformly sample 10 clips from a video along its temporal axis.
For each clip, we resize the shorter spatial side to 256 pixels and take 3 crops of 256\x256 along the longer side to cover the spatial dimensions. Video-level predictions are computed by averaging softmax scores.
We report the actual \emph{inference-time} computation as in \cite{slowfast}, by listing the FLOPs per spacetime ``view" of spatial size 256$^2$ (temporal clip with spatial crop) at inference \emph{and} the number of views (\ie 30 for 10 temporal clips each with 3 spatial crops). 

Kinetics, EPIC, Charades details are in \ref{sec:kinetics_details}, \ref{sec:epic_details}, \& \ref{sec:charades_details}.

\subsection{Main Results}

\paragraph{EPIC-Kitchens.} 

We compare to state-of-the-art methods on EPIC in Table~\ref{table:epic}. 
First, AVSlowFast improves SlowFast by \textbf{+2.9} / \textbf{+4.3 }/ \textbf{+2.3} top-1 accuracy for verb / noun / action, which highlights the benefits of audio in egocentric video recognition. 
\begin{table}[tb]
	\vspace{-25pt}
	\hspace*{-5pt}
	\centering
	\resizebox{\columnwidth}{!}{
		\tablestyle{1.8pt}{1.05}
		\begin{tabular}{@{}l|x{22}x{22}x{22}x{22}x{22}x{22}@{}}
			\multicolumn{1}{c}{} 		& \multicolumn{2}{c}{verbs} & \multicolumn{2}{c}{nouns} & \multicolumn{2}{c}{actions}\\
			\cline{2-3}
			\cline{4-5}
			\cline{6-7}
			\multicolumn{1}{c|}{model} 			& top-1 & top-5 & top-1 & top-5 & top-1 & top-5\\
			\shline
			\textbf{validation}\\
			{\quad 3D CNN}~\cite{wu-lfb2019} & 49.8 & 80.6 & 26.1 & 51.3 & 19.0 & 37.8\\
			{\quad LFB}~\cite{wu-lfb2019} & 52.6  & 81.2 & \textbf{31.8} & 56.8 & 22.8 & 41.1\\
			{\quad SlowFast}~\cite{slowfast} &  55.8 &  83.1 &  27.4 &  52.1 &  21.9 &  39.7\\
			{\quad \bf AVSlowFast} &  \textbf{58.7} &  \textbf{83.6} &  {31.7} &  \textbf{58.4} &  \textbf{24.2} &  \textbf{43.6}\\
			\quad\quad\ \ $\Delta$ & \emph{+2.9} & \emph{+0.5} & \emph{+4.3} & \emph{+6.3} & \emph{+2.3} & \emph{+3.9}\\
			\hline
			\textbf{test s1 (seen)}\\
			{\quad HF-TSN}~\cite{sudhakaran-arxiv2019} &  57.6 &  87.8 &   39.9 &  65.4 &   28.1 &  48.6 \\            
			{\quad RU-LSTM}~\cite{furnari2019would} & 	56.9  &  85.7 &   43.1  &67.1  &  33.1  &  55.3  \\
			{\quad FBK-HUPBA}~\cite{sudhakaran-fbk2019} &  63.3 &  89.0 &  44.8 &  69.9 &  35.5 &  57.2 \\
			{\quad LFB}~\cite{wu-lfb2019} &  60.0 &  88.4 &  45.0 &  \textbf{71.8} &  32.7 &  55.3\\
			{\quad EPIC-Fusion}~\cite{kazakos-epicfusion2019} &  64.8 &  \textbf{90.7} &   46.0 &   71.3 &   34.8 &  56.7\\
			{\quad \bf AVSlowFast} &  \textbf{65.7} &  89.5 &  \textbf{46.4} &  71.7 &  \textbf{35.9} &  \textbf{57.8}\\
			\hline
			\textbf{test s2 (unseen)}\\
			{\quad HF-TSN}~\cite{sudhakaran-arxiv2019} &  42.4 &  75.8 &   25.2 &  49.0 &  16.9 &  33.3 \\
			{\quad RU-LSTM}~\cite{furnari2019would} & 43.7   &  73.3 &  26.8  &48.3  &  19.5  &  37.2  \\
			{\quad FBK-HUPBA}~\cite{sudhakaran-fbk2019} &  49.4 &  77.5 &  27.1 &  52.0 &  20.3 &  37.6 \\
			{\quad LFB~\cite{wu-lfb2019}} &  50.9 &  77.6 &  31.5 &  57.8 &  21.2 &  39.4\\
			{\quad EPIC-Fusion}~\cite{kazakos-epicfusion2019} &  52.7 &  79.9 &   27.9 &   53.8 &   19.1 &  36.5\\
			{\quad \bf AVSlowFast} &  \textbf{55.8} &  \textbf{81.7} &  \textbf{32.7} &  \textbf{58.9} &  \textbf{24.0} &  \textbf{43.2}\\
			\quad\quad\ \ $\Delta$ & \emph{+3.1} & \emph{+1.8} & \emph{+4.8} & \emph{+4.9} & \emph{+2.3} & \emph{+6.7}\\
		\end{tabular}
	}
	\vspace{0.1em}
	\caption{\textbf{EPIC-Kitchens validation and test results.} Models pretrain on  Kinetics \cite{slowfast,kazakos-epicfusion2019,wu-lfb2019} or ImageNet \cite{sudhakaran-arxiv2019,sudhakaran-fbk2019,furnari2019would}. SlowFast backbones: $T$\x$\tau$ $=$ 8\x8, R101. AVSlowFast shows strong margins ($\Delta$) over SlowFast and previous state-of-the-art~\cite{kazakos-epicfusion2019}.
	}\label{table:epic}
	\vspace{-10pt}
\end{table}
Second, as a system-level comparison,  AVSlowFast exhibits higher performance
 in all three categories (verb/noun/action) and two test sets (seen/unseen) vs.~state-of-the-art~\cite{kazakos-epicfusion2019} under Kinetics-400 pretraining.

\begin{table}[t!]
	\vspace{-25pt}
	\centering
	\small
	\hspace*{-7pt}
	\tablestyle{1.8pt}{1.05}
	\begin{tabular}{l|c|c|c|c|c|c}
		\multicolumn{1}{c|}{model} & \multicolumn{1}{c|}{inputs}  & \multicolumn{1}{c|}{pretrain} &  top-1  & top-5  & KS &  GFLOPs\x views  \\
		\shline
		
		R(2+1)D~\cite{tran-cvpr2018} & V  & - & 72.0 &  90.0 &  \multirow{12}{*}{N/A}  & 152~\x~115 
		\\
		TS R(2+1)D~\cite{tran-cvpr2018} & V+F & - & 73.9 &  90.9 &  & 304~\x~115
		\\
		ECO~\cite{zolfaghari-eccv2018} & V &  - &  70.0 &  89.4 &  & N/A~\x~N/A 
		\\
		ip-CSN-152  \cite{tran-csn2019}  & V &  -&  77.8 & 92.8 &  & 109~\x~30 
		\\
		S3D~\cite{xie-eccv2018}  & V & - & 69.4 & 89.1 &  & 66.4~\x~N/A  \\
		I3D~\cite{carreira-i3d2017} & V & \checkmark & 72.1 &  90.3 & & 108~\x~N/A \\
		TS I3D~\cite{carreira-i3d2017} & V+F & \checkmark & 75.7 &  92.0 &  & 216~\x~N/A \\
		TS I3D~\cite{carreira-i3d2017} & V+F & - &  71.6 & 90.0 &  & 216~\x~N/A\\
		Nonlocal~\cite{wang-nonlocal2018}, R101 & V & \checkmark & 77.7 & 93.3 &  & 359~\x~30
		\\
		S3D-G \cite{xie-eccv2018} & V & \checkmark & 74.9 &  92.0 &  & 71.4~\x~N/A \\
		TS S3D-G \cite{xie-eccv2018} & V+F & \checkmark & 77.2 &  93.0 &  & 143~\x~N/A \\
		3-stream SATT~\cite{bian-arxiv2017} & A+V+F & \checkmark & 77.7 & 93.2 &  & N/A~\x~N/A
		\\
		\hline
		SlowFast, R50 \cite{slowfast}& V & - & 75.6 & 92.0 & 80.5 & 36~\x~30 
		\\
		\textbf{AVSlowFast}, R50 & A+V & - &  77.0 & 92.7 & 83.7 & 40~\x~30 
		\\
		SlowFast, R101 \cite{slowfast}& V & - & 77.9 & 93.2 & 82.7 & 106~\x~30 
		\\
		\textbf{AVSlowFast}, R101 & A+V & - &  \textbf{78.8} & \textbf{93.6} & \textbf{85.0} & 129~\x~30 
		\\
	\end{tabular}
	\hspace{-0.8em}
	\caption{\textbf{AVSlowFast results on Kinetics.} AVSlowFast and \mbox{SlowFast} instantiations are with $T$\x$\tau$ $=$ 4\x16 and $T$\x$\tau$ $=$ 8\x8 inputs for R50/R101 backbones, without NL blocks. ``TS'' indicates Two-Stream. ``KS'' refers to top-1 accuracy on Kinetics-Sounds dataset~\cite{arandjelovic-iccv2017}. ``pretrain'' refers to ImageNet pretraining.
	} 
	\label{table:kinetics}
	\vspace{-5pt}
\end{table}

\begin{table}[h!]  
	\centering
	\footnotesize
	\tablestyle{1.8pt}{1.05}
	\begin{tabular}{l|x{68}|x{22}|c}
		\multicolumn{1}{c|}{model} &  \multicolumn{1}{c|}{pretrain} &  mAP    & \scriptsize GFLOPs\x views    \\ 
		\shline
		Nonlocal, R101~\cite{wang-nonlocal2018} &  {\scriptsize ImageNet+Kinetics}  &   37.5 & 544~\x~30  \\
		STRG, R101+NL~\cite{wang-eccv2018} &  {\scriptsize ImageNet+Kinetics} &   39.7 & 630~\x~30  \\
		Timeception \cite{hussein-timeception2019} & {\scriptsize Kinetics-400} & 41.1 &  N/A\x N/A  \\
		LFB, +NL \cite{wu-lfb2019} & {\scriptsize Kinetics-400}  & 42.5 & 529 ~\x~30   \\
		\hline
		SlowFast  & {\scriptsize Kinetics-400} & 42.5 &  234~\x~30  \\
		\textbf{AVSlowFast}  & {\scriptsize Kinetics-400} & \textbf{43.7} &  278~\x~30  \\
	\end{tabular}
	\vspace{.5em}
	\caption{\textbf{Comparison with the state-of-the-art on Charades}. AV/SlowFast is with R101+NL backbone and 16\x8 sampling.
	} 	
	\label{table:charades}
	\vspace{-10pt}
\end{table}

\begin{table*}[t!]\centering 
	\vspace{-20pt}
	\captionsetup[subfloat]{captionskip=2pt}
	\captionsetup[subffloat]{justification=centering}
	\subfloat[\textbf{Audiovisual fusion}.\label{table:fusionconnection}]{
		\tablestyle{2pt}{1.05}
		\hspace*{-5mm}
		\begin{tabular}{l | c c c c}
			\multicolumn{1}{c|}{connection} & top-1 & top-5 & GFLOPs 
			\vspace{0.05cm} 
			\\
			\shline
			A$\rightarrow$F$\rightarrow$S & 75.3 & 91.8 & 51.4 
			\\
			A$\rightarrow$FS & 77.0 & 92.7 & 39.8 
			\\
			AV Nonlocal & \textbf{77.2} & 92.9 & 39.9 
			\\
			\multicolumn{3}{c}{} 
	\end{tabular}}\hspace{0mm}
	\subfloat[\textbf{Audio channels $\beta_A$}.\label{table:betaa}]{
		\tablestyle{2pt}{1.05}
		\begin{tabular}{c | c c c}
			$\beta_A$ & top-1 & top-5 & GFLOPs 
			\vspace{0.05cm} 
			\\
			\shline
			1/8 & 76.0 & 92.5 & 36.0 
			\\
			1/4 & 76.6 & 92.7 & 36.8 
			\\
			1/2 & \textbf{77.0} & 92.7 & 39.8 
			\\
			1 & 75.9 & 92.4 & 51.9 
	\end{tabular}}\hspace{0mm}
	\subfloat[\textbf{DropPathway rate $P_d$}. \label{table:droppathway}]{
		\tablestyle{2pt}{1.05}
		\begin{tabular}{x{30} | x{35} x{35}}
			${P}_{d}$ & top-1 & top-5
			\vspace{0.05cm} 
			\\
			\shline
			- & 75.2 & 91.8  
			\\
			0.2 & 76.0 & 92.5 
			\\
			0.5 & 76.7 & 92.7
			\\
			0.8 & \textbf{77.0} & 92.7
	\end{tabular}}\hspace{0mm}
	\subfloat[\textbf{AV synchronization}. \label{table:avs}]{
		\tablestyle{2pt}{1.05}
		\begin{tabular}{x{30} | x{35} x{35}}
			\multicolumn{1}{c|}{AVS} & top-1 & top-5 
			\vspace{0.02cm} 
			\\
			\shline
			- & 76.4 & 92.5 
			\\
			res$_{5}$ & 76.7 & 92.8 
			\\
			res$_{4,5}$ & 76.9 & 92.9 
			\\
			res$_{3,4,5}$ & \textbf{77.0} & 92.7 
		\end{tabular}
		\hspace*{-5mm}
	}
	\caption{\textbf{Ablations on AVSlowFast design} on Kinetics-400. We show top-1/5 classification accuracy (\%), and computational complexity measured in GFLOPs for a  single clip input of spatial size 256$^2$. Backbone: 4\x 16, R-50. }
	\label{tab:ablations}
	 \vspace{-12pt}
\end{table*}

Comparing to LFB~\cite{wu-lfb2019}, which uses an object detector to localize objects, AVSlowFast achieves similar performance for nouns (objects) on both the seen and unseen test sets, whereas SlowFast \textit{without audio} is largely lagging behind (-4.4\% \vs LFB on val noun), which is intuitive as sound can be beneficial for recognizing objects.

We observe large performance gains over previous best \cite{kazakos-epicfusion2019} (which utilizes rgb, audio and flow) on the unseen split (\ie, novel scenes) of the test set (\textbf{+3.1} / \textbf{+4.8}/ \textbf{+2.3} for verb / noun / action), showing \mbox{AVSlowFasts}' strength on test data.  

\paragraph{Kinetics.} 

Table~\ref{table:kinetics} shows a comparison on the well-established Kinetics dataset.  Comparing AVSlowFast with SlowFast shows a margin of 1.4\% top-1 for R50 and 0.9\% top-1 accuracy for R101, given the same network backbone and input size.
This demonstrates the effectiveness of the audio stream despite its modest cost of only $\approx$10\%$-$20\% of the overall computation. Comparatively, going deeper from R50 to R101 increases computation by 194\%. 

On a system-level, AVSlowFast compares favorably to existing methods that utilize various modalities, \ie, audio (A), visual frames (V) and optical flow (F). Adding optical flow streams brings roughly similar gains as audio but \emph{doubles} computation (TS in Table~\ref{table:kinetics}), not counting optical flow computation; by contrast, audio processing is lightweight (\eg 11\% computation overhead for AVSlowFast, R50). Further, AVSlowFast does not rely on pretraining and is competitive with multi-modal approaches that pretrain individual modality streams (\checkmark). 

As Kinetics is a visual-heavy dataset (for many classes \eg ``\textit{writing}'' audio is not useful), to better study audiovisual learning, ``Kinetics-Sounds'' \cite{arandjelovic-iccv2017} is as a subset of 34 classes potentially manifested both visually and aurally. We test on Kinetics-Sounds in the ``KS'' column of Table~\ref{table:kinetics}. The gain from SlowFast to AVSlowFast doubled -- for R50/R101 with +3.2\%/+2.3\%, showing the potential on relevant data. Further Kinetics results on standalone Audio-only classification and class-level analysis are in \ref{sec:results_audio_only} and \ref{sec:results_analysis}.

\paragraph{Charades.} 
We test the effectiveness of AVSlowFast on videos of longer range activities on Charades in Table~\ref{table:charades}.  We observe that audio can  facilitate recognition (+1.2\% over a strong  SlowFast baseline) and we achieve state-of-the-art performance under Kinetics-400 pre-training. 

\paragraph{Discussion.} Overall, our experiments on action classification indicate that, on standard, visually created datasets for classification, a consistent improvement over very strong visual baselines can be achieved by modeling audio with AVSlowFast. For some cases improvements are exceptionally high (\eg EPIC) and in some lower (\eg Charades), and all results suggest that with AVSlowFast, audio can serve as an \emph{economical} modality that supplements visual input.

\subsection{Ablation Studies}\label{sec:ablation}
We ablate of our approach on Kinetics as it represents the largest unconstrained dataset for human action recognition.

\begin{table}[h!]
	 \vspace{-5pt}
	\centering
	\footnotesize
	\begin{tabular}{l | c c c c}
		\multicolumn{1}{c|}{fusion stage} & top-1 & top-5 & GFLOPs 
		\\
		\shline
		\demph{SlowFast} & \demph{75.6} & \demph{92.0} & \demph{36.1} 
		\\
		\demph{SlowFast+Audio} & \demph{76.1} & \demph{92.0} & \demph{-}
		\\
		pool$_5$ & 75.4 & 92.0 & 38.4 
		\\
		res$_4$ + pool$_5$  & 76.5 & 92.6 & 39.1 
		\\
		res$_{3,4}$ + pool$_5$ & \textbf{77.0} & \textbf{92.7} & 39.8 
		\\
		res$_{2,3,4}$ + pool$_5$ & 75.8 &  92.4 & 40.2 
	\end{tabular}
	\vspace{0.1cm}
	
	\caption{\textbf{Effects of hierarchical fusion.} Backbone: 4\x 16, R-50. }
	\label{table:multilevelfusion}
	\vspace{-10pt}
\end{table}

\paragraph{Hierarchical fusion.}
We study the effectiveness of fusion in Table~\ref{table:multilevelfusion}.
The first interesting phenomenon is that direct ensembling (late-fusion) of audio/visual models produces {only} modest gains (76.1\% vs 75.6\%), whereas joint training with late-fusion (``pool$_5$'') does not help (75.6\% $\rightarrow$ 75.4\%).

For hierarchical, multi-level fusion,  Table~\ref{table:multilevelfusion} shows it is beneficial to fuse audio and visual features at multiple levels. Specifically, we found that recognition accuracy steadily increases from 75.4\% to 77.0\% when we increase the number of fusion connections from one (\ie, only concatenating pool5 outputs) to three (res$_{3,4}$ + pool$_5$) where it peaks. Adding another lateral connection at res$_2$ decreases accuracy. This suggests that it is beneficial to start fusing audio and visual features from intermediate levels (res$_3$) all the way to the top of the network.  We hypothesize that this is because audio facilitates the formation of visual concepts, but only when features mature to intermediate concepts that are generalizable across modalities (\eg local edges do not have a general sound pattern). 

\paragraph{Lateral connections.} We ablate the the effect of different types of lateral connections between audio and visual pathways in Table~\ref{table:fusionconnection}. First, A$\rightarrow$F$\rightarrow$S, which enforces strong temporal alignment between audio and visual streams, produces lower classification accuracy compared to A$\rightarrow$FS, which relaxes the requirement on alignment. This {is consistent with findings in~\cite{kazakos-epicfusion2019}} that it is beneficial to have tolerance on alignment between the modalities, since class-level audio signals might happen out-of-sync to visual frames (\eg, when shooting 3 pointers in basketball, the net-touching sound only comes after the action finishes). Finally, the straightforward A$\rightarrow$FS connection performs similarly to the more complex AV Nonlocal \cite{wang-nonlocal2018} fusion (77.0\% vs 77.2\%). We use A$\rightarrow$FS as our default lateral connection for its good performance and simplicity. 

\paragraph{Audio pathway capacity.} We study the impact of the number of channels of the Audio pathway  ($\beta_A$) in Table~\ref{table:betaa}. As expected, when we increase the number of channels (\eg, increasing $\beta_A$ from 1/8 to 1/2, which is the ratio between Audio and Slow pathway's channels), accuracy improves at the cost of increased computation. However, performance starts to degrade when we further increase it to 1, likely due to overfitting. We use $\beta_A = 1/2$ across all our experiments.   

\paragraph{DropPathway.} We apply Audio pathway dropping to adjust the incompatibility of learning speed across modalities. Here we conduct ablative experiments to study the effects of different drop rates ${P}_{d}$. The results are shown in Table~\ref{table:droppathway}. As shown in the table, a high value of $P_{d}$ (0.5 or 0.8) is required to slow down the Audio pathway when training audio and visual pathways jointly. If we train AVSlowFast without \mbox{DropPathway} (``-''), the accuracy degrades to be even worse than visual-only models (75.2\% vs 75.6\%). This is because the Audio pathway learns too fast and starts to dominate the visual feature learning. The gain from \mbox{75.2\% $\rightarrow$ 77.0\%} reflects the full impact of DropPathway. 

\paragraph{Hierarchical audiovisual synchronization.} We study the effectiveness of hierarchical audiovisual synchronization in Table~\ref{table:avs}. We use AVSlowFast with and without AVS, and vary the layers for multiple losses. We observe that adding AVS as an auxiliary task is beneficial (+0.6\% gain). Furthermore, having synchronization loss at multiple levels slightly increases the performance (without extra inference cost). 
This suggests that it is beneficial to have a feature representation that is generalizable across audio and visual modalities and {hierarchical AVS} could facilitate producing such. 

\section{Experiments: AVA Action Detection}\label{sec:ava}
\vspace{-5pt}

In addition to the action classification tasks, we also apply AVSlowFast models on action detection which requires both localizing and recognizing actions.

\paragraph{Dataset.} 
	The AVA dataset \cite{gu-ava2018} focuses on spatiotemporal localization of human actions. Spatiotemporal labels are provided for one frame per second, with people annotated with a bounding box and (possibly multiple) actions. 
There are 211k training and 57k validation video segments. We follow the standard protocol  \cite{gu-ava2018} of evaluating on 60 classes.
The metric is mean Average Precision (mAP) over 60 classes, using a frame-level IoU threshold of 0.5. 

\paragraph{Detection architecture.}
We follow the detection architecture introduced in~\cite{slowfast}, which is adapted from Faster R-CNN~\cite{ren-nips2015} for video. 
During training, the input to our audiovisual detector is $\alpha_FT~$ RGB frames sampled with temporal stride $~\tau$ and spatial size 224\x224, to SlowFast pathways, and the corresponding log-mel-spectrogram covering this time window to Audio pathway. During testing, the backbone feature is computed fully convolutionally with RGB frame of shorter side being 256 pixels \cite{slowfast}, as is standard in Faster R-CNN \cite{ren-nips2015}. 

For details on architecture, training and inference, please refer to appendix~\ref{sec:ava_details}.

\begin{table}[t!]
	\vspace{-10pt}
	\centering
	\tablestyle{2.5pt}{1.05}
	\begin{tabular}{l|c|c|c|c|c}
		\multicolumn{1}{c|}{model} & inputs & AVA & \multicolumn{1}{c|}{pretrain} &  val mAP & GFLOPs \\ 
		\shline
		I3D~\cite{gu-ava2018} & V+F & \multirow{11}{*}{v2.1}  & \multirow{11}{*}{K400}   & 15.6 & N/A\\
		ACRN,  S3D~\cite{sun-eccv2018} & V+F & &  & 17.4 & N/A\\
		ATR, R50+NL~\cite{jiang-cvpr2018}    & V+F & &  & 21.7 & N/A\\
		9-model ensemble~\cite{jiang-cvpr2018}  & V+F & &  & 25.6 & N/A\\
		I3D+Transformer \cite{girdhar-transformer2019} \quad\ &  V & &   &  25.0 & N/A \\
		
		LFB, + NL R50 \cite{wu-lfb2019} &   V& & & 25.8 &  N/A   \\ 
		LFB, + NL R101 \cite{wu-lfb2019} & V & & & 26.8 & N/A \\ \hline
		
		SlowFast 4\x16, R50 & V &  &    & 24.3 & 65.7 \\ 
		\textbf{AVSlowFast} 4\x16, R50 & A+V &  &  & 25.4 & 67.1 \\ 
		SlowFast 8\x8, R101 & V &  &    & 26.3 & 184 \\ 
		\textbf{AVSlowFast} 8\x8, R101 & A+V &  &  & \textbf{27.8} & 210 \\ 
		\hline
		SlowFast 4\x16, R50 & V & \multirow{4}{*}{v2.2} & \multirow{4}{*}{K400}     & 24.7 & 65.7 \\ 
		\textbf{AVSlowFast} 4\x16, R50 & A+V & &  & 25.9 &  67.1 \\
		SlowFast 8\x8, R101 & V &  &    & 27.4 & 184\\ 
		\textbf{AVSlowFast} 8\x8, R101 & A+V &  &  & \textbf{28.6} & 210 
	\end{tabular} 
	\vspace{0.1cm}
	\caption{\small \textbf{Comparison on AVA detection}. AVSlowFast and SlowFast use 8\x8 inputs. For R101, both use NL blocks~\cite{wang-nonlocal2018}.}
	\label{table:ava}
	\vspace{-1.0em}
\end{table}

\paragraph{Results.} 
We compare to  several other existing methods in Table~\ref{table:ava}. AVSlowFast, with both R50 and R101 backbones, outperforms SlowFast with a consistent margin of \app1.2\%, and only increases FLOPs\footnote{We report FLOPs for fully-convolutional inference of a clip with 256\x320  spatial size for SlowFast and AVSlowFast models, full test-time computational cost for these models is directly proportional to this. } slightly, \eg for R50  \textit{by only 2\%}, whereas going from SlowFast R50 to R101 (without audio) increases computation significantly by 180\%.

Interestingly, the ActivityNet Challenge 2018 \cite{ghanem-activitynet2018} hosted a separate track for multiple modalities, but no team could achieve gains using audio information on AVA data. Our result shows, for the first time, that audio can be beneficial for action detection, where spatiotemporal localization is required, even with low computation overhead of just 2\%.

For system-level comparison to other approaches, Table~\ref{table:ava} shows that AVSlowFast achieves state-of-the-art performance on AVA under Kinetics-400  pretraining. 

For comparisons with future work, we show results on the newer v2.2 of AVA, which provides updated annotations. We see consistent results as for v2.1. As for per-class results, we found classes like [``{\it swim}'' +30.2\%], [``{\it dance}'' +10.0\%], [``{\it shoot}'' +8.6\%], and [``{\it hit (an object)}'' +7.6\%] has the largest gain from audio; please see  appendix~\ref{sec:results_analysis} and \figref{fig:ava} for more details.

\section{Experiments: Self-supervised Learning} \label{sec:ssl}

\begin{table}[t!]
	\vspace{-10pt}
	\centering
	\footnotesize
	\tablestyle{2.5pt}{1.05}
	\begin{tabular}{l | c c c c c c}
		\multicolumn{1}{c|}{method} & inputs & \#param & FLOPs & pretrain & UCF & HMDB 
		\\
		\shline
		Shuffle\&Learn~\cite{misra-eccv2016,sun-cbt2019} & V & 58.3M  & N/A & K600 & 26.5 & 12.6
		\\
		3D-RotNet~\cite{jing-arxiv2018,sun-cbt2019} & V &  33.6M  & N/A &   K600 & 47.7 & 24.8
		\\
		CBT~\cite{sun-cbt2019} & V & N/A & N/A & K600 & 54.0 & 29.5
		\\ 
		\hline
		\textbf{AVSlowFast} & A+V & 38.5M  & 63.4G &  K400 & \textbf{77.4} & \textbf{42.2}
	\end{tabular}
	\vspace{2pt}
	\caption{\textbf{Comparison using the linear classification protocol}.   	
		We only train the last \textit{fc} layer after SSL pretraining AVSlowFast features on Kinetics. Top-1 accuracy averaged over three splits is reported. Backbone: $T$ \x~$\tau$ = 8 \x~8, R50.
	}
	\label{table:ssl}
	\vspace{-5pt}
\end{table}

To further study the generalization of AVSlowFast models, we apply it to self-supervised learning (SSL). 
The goal here is not to propose a new SSL pretraining task. Instead, we are interested in how well a self-supervised video representation can be learned with AVSlowFast using existing tasks. We use the audiovisual synchronization~\cite{arandjelovic-iccv2017,korbar-scsampler2019,owens-eccv2018} and image rotation prediction~\cite{gidaris-iclr2018} (0$\degree$, 90$\degree$, 180$\degree$, 270$\degree$; as a four-way softmax-classification) losses as pretraining tasks.  With the learned {AVSlowFast weights}, we then re-train the last \textit{fc} layer of AVSlowFast on UCF101~\cite{soomro-ucf2012} and HMDB51~\cite{kuehne-hmdb2011} following standard practice to evaluate the SSL feature representation. Table~\ref{table:ssl} lists the results. Using off-the-shelf pretext tasks, our smallest AVSlowFast, R50 model compares favorably to state-of-the-art SSL approaches on both datasets, with an absolute margin of \textbf{+23.4} and \textbf{+12.7} top-1 accuracy over previous best  CBT~\cite{sun-cbt2019}. This is highlighting the strength of the architecture, and the features learned by AVSlowFast.  For more details and results, please refer to appendix \ref{sec:results_ssl}.

\section{Conclusion}
This work has presented AVSlowFast \mbox{Networks}, an architecture for integrated audiovisual perception. We show the effectiveness of the AVSlowFast representation with state-of-the-art performance on six datasets for video action classification, detection, and self-supervised learning tasks. We hope that AVSlowFast, as a unified audiovisual backbone, will foster further research in video understanding.

\newcount\cvprrulercount
\appendix
\section{Appendix}

\setcounter{table}{0}
\renewcommand{\thetable}{A.\arabic{table}}	

\setcounter{figure}{0}
\renewcommand{\thefigure}{A.\arabic{figure}}	
\subsection{Results: Self-supervised Learning}  \label{sec:results_ssl}
\begin{table}[h!]
	\vspace{-0.2cm}
	\centering
	\footnotesize
	\tablestyle{1.4pt}{1.05}
	\begin{tabular}{l | c c c c c c}
		\multicolumn{1}{c|}{method} & inputs & \#param & FLOPs & pretrain & UCF & HMDB 
		\\
		\shline
		Shuffle\&Learn~\cite{misra-eccv2016,sun-cbt2019} & V & 58.3M  & N/A & K600 & 26.5 & 12.6
		\\
		3D-RotNet~\cite{jing-arxiv2018,sun-cbt2019} & V &  33.6M  & N/A &   K600 & 47.7 & 24.8
		\\
		CBT~\cite{sun-cbt2019} & V & N/A & N/A & K600 & 54.0 & 29.5
		\\ 
		\hline
		\textbf{AVSlowFast} 4\x16 & A+V & 38.5M  & 36.2G &  K400 & 76.8 & 41.0
		\\
		\textbf{AVSlowFast} 8\x8 & A+V & 38.5M  & 63.4G &  K400 & 77.4 & 42.2
		\\ 
		\textbf{AVSlowFast} 16\x4 & A+V & 38.5M  & 117.9G &  K400 & \textbf{77.4} & \textbf{44.1}
		\\
		\shline
		\textbf{ablation (split1)}
		\\
		\quad SlowFast 4\x16 (ROT) & V & 33.0M  & 34.2G &  K400 & 71.9 & \textbf{42.0}
		\\
		\quad AVSlowFast 4\x16 (AVS) & A+V & 38.5M  & 36.2G &  K400 & 73.2 & 39.5
		\\
		\quad \textbf{AVSlowFast} 4\x16 & A+V & 38.5M  & 36.2G &  K400 & \textbf{77.0} & 40.2
	\end{tabular}
	\vspace{0.1pt}
	\caption{\textbf{Comparison using the linear classification protocol}.   	
		We only train the the last \textit{fc} layer after self-supervised pretraining on Kinetics-400 (abbreviated as K400). Top-1 accuracy averaged over three splits is reported when comparing to previous work (top), results on split1 is used for ablation (bottom). All SlowFast models use use R50 backbones with $T$ \x~$\tau$  sampling. 
	}
	\label{table:sslftfc}
\end{table}

\begin{table}[t!]
	\centering
	\footnotesize
	\tablestyle{1.2pt}{1.05}
	\begin{tabular}{l | c c c c c}
		\multicolumn{1}{c|}{method} & inputs & \#param & pretrain & UCF101 & HMDB51 
		\\
		\shline
		Shuffle \& Learn~\cite{misra-eccv2016} & V & 58.3M & UCF/HMDB & 50.2 & 18.1
		\\
		OPN~\cite{lee-iccv2017} & V & 8.6M & UCF/HMDB & 59.8 & 23.8
		\\
		O3N~\cite{fernando-cvpr2017} & V &  N/A &  Kinetics-400 & 60.3 & 32.5 
		\\
		3D-RotNet~\cite{jing-arxiv2018} & V &  33.6M &   Kinetics-400 & 62.9 & 33.7
		\\
		3D-ST-Puzzle~\cite{kim-aaai2019} & V & 33.6M &   Kinetics-400 & 65.8 & 33.7
		\\
		DPC~\cite{han-dpc2019} & V &  32.6M &   Kinetics-400 & 75.7 & 35.7
		\\
		CBT~\cite{sun-cbt2019} & V & N/A & Kinetics-600 & 79.5 &  44.6
		\\
		Multisensory~\cite{owens-eccv2018} & A+V &  N/A &  Kinetics-400 & 82.1 & N/A 
		\\
		AVTS~\cite{korbar-nips2018} & A+V &  N/A &  Kinetics-400 & 85.8 & \textbf{56.9}
		\\ 
		\hline
		VGG-M motion~\cite{simonyan-nips2014,feichtenhofer-cvpr2016} & V &    90.7M & - & 83.7 & 54.6
		\\ 
		\hline
		\textbf{AVSlowFast} & A+V & 38.5M &  Kinetics-400 & \textbf{87.0} & 54.6
		
	\end{tabular}
	\vspace{3pt}
	\caption{\textbf{Comparison for Training all layers.} Results using the popular protocol of fine-tuning all layers after self-supervised pretraining.  Top-1 accuracy averaged over three splits is reported.   We use AVSlowFast 16\x4, R50 for this experiment. While this protocol has been used in the past, we think it is suboptimal for evaluation of self-supervised representations, as the training of all layers can significantly impact performance; \eg an AlexNet-like VGG-M motion stream \cite{simonyan-nips2014,feichtenhofer-cvpr2016} can perform among state-oft-the-art self-supervised approaches, {without any pretraining}. }
	\label{table:sslftall}
	\vspace{-15pt}
\end{table}

In this section, we provide more results and detailed analysis on self-supervised learning using AVSlowFast. 
Training schedule and details are provided in \S\ref{sec:ssl_details}.

First, we pretrain AVSlowFast with self-supervised objectives of audiovisual synchronization~\cite{arandjelovic-iccv2017,korbar-scsampler2019,owens-eccv2018} (AVS) and image rotation prediction~\cite{gidaris-iclr2018} (ROT) on Kinetics-400. Then, following the standard linear classification protocol used for image recognition tasks~\cite{he-moco2019}, we use the pretrained network as a fixed, \textit{frozen} feature extractor and train a linear classifier on top of the self-supervisedly learned features. 
In Table~\ref{table:sslftfc} (top), we compare to previous work that follows the \textit{same protocol}. We note this is the same experiment as in \tblref{table:ssl}, but with additional ablations on our models.  
The results indicate that features learned by AVSlowFast are significantly better than baselines including the recently introduced CBT method~\cite{sun-cbt2019} (+23.4\% for UCF101 and +14.6\% for HMDB51), which also uses ROT as well as a contrastive bidirectional transformer (CBT) loss by pretraining on the larger Kinetics-600. 

In addition, we also ablate the contribution of individual tasks of AVS and ROT in Table~\ref{table:sslftfc} (bottom). On UCF101, SlowFast/AVSlowFast trained under either ROT or AVS objective show strong individual performance, while the combination of them perform the best. Whereas on the smaller HMDB51, all three variants of our method perform similarly well and audio seems less important. 

Another aspect is that, although many previous approaches on self-supervised feature learning focus on reporting number of parameters, the FLOPs are another important factor to consider -- as shown in Table~\ref{table:sslftfc} (top), the performance keeps increasing when we take higher temporal resolution clips by varying $T$\x$\tau$ (\ie larger FLOPs), even though model \textit{parameters remain identical}. 

Although we think the linear classification protocol serves as a better method to evaluate self-supervised feature learning (as features are frozen and therefore less sensitive to hyper-parameter settings such as learning schedule and regularization, especially when these datasets are relatively small), we also evaluate by fine-tuning all layers of AVSlowFast on the target datasets to compare to a larger corpus of previous work on self-supervised feature learning. Table~\ref{table:sslftall} shows that AVSlowFast achieves competitive performance comparing to prior work under this setting.
When using this protocol, we believe it is reasonable to also consider methods that train multiple layers on UCF/HMDB from scratch, such as optical-flow based motion streams~\cite{simonyan-nips2014,feichtenhofer-cvpr2016}. It is interesting that this stream, despite being an AlexNet-like model~\cite{Chatfield2014a},  is comparable or better, than many newer models, pretrained on (the large) Kinetics-400 using self-supervised learning techniques. 	

\begin{figure*}[t]
	\centering
	\hspace*{-19pt}
	\includegraphics[width=1.1\textwidth]{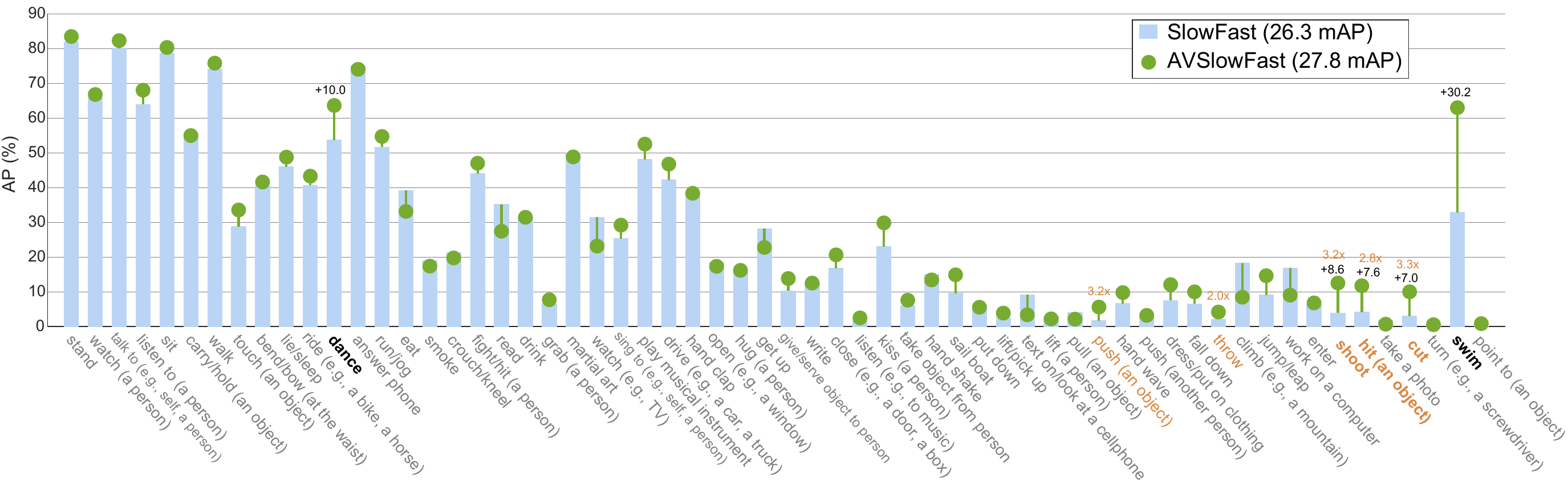}
	\caption{\textbf{AVA per-class average precision}.   
		AVSlowFast (27.8 mAP) \vs its SlowFast counterpart (26.3 mAP). The highlighted categories are the 5 highest absolute increases (\textbf{bold}) and top 5 relative increases over SlowFast ({\color{orange}{\textbf{orange}}}).  Best viewed in color with zoom. 
	}
	\label{fig:ava}
\end{figure*}

\subsection{Results: Audio-only Classification} \label{sec:results_audio_only}

To understand the effectiveness of our Audio pathway, we evaluate it in terms of Audio-only classification accuracy on Kinetics (in addition to Kinetics-400, we also train and evaluate on Kinetics-600 to be comparable to methods that use this data in challenges \cite{ghanem-activitynet2018}). In Table~\ref{table:audioonly}, we compare our Audio-only network to several other audio models. We observe that our Audio-only model performs better than existing methods by solid margins (+3.3\% top-1 accuracy on Kinetics-600 and +3.2\% on Kinetics-400, compared to best-performing methods), which demonstrates the effectiveness of our Audio pathway design. Note also that unlike some other methods in Table~\ref{table:audioonly}, we train our audio network from scratch on Kinetics, without any pretraining. 

\begin{table}[h]
	\centering
	\footnotesize
	\tablestyle{1.9pt}{1.05}
	\begin{tabular}{l | c | c | c | c | c c}
		\multicolumn{1}{c|}{model} & dataset & pretrain & top-1 & top-5 & GFLOPs
		\\
		\shline
		VGG*~\cite{hershey-vggish2017} & Kinetics-600 & Kinetics-400 & 23.0 & N/A & N/A 
		\\
		SE-ResNext~\cite{ghanem-activitynet2018} & Kinetics-600 & ImageNet &  21.3 & 38.7 & N/A 
		\\
		Inception-ResNet~\cite{ghanem-activitynet2018} & Kinetics-600 & ImageNet & 23.2 & N/A & N/A 
		\\
		\textbf{Audio-only} (ours) & Kinetics-600 & - & \textbf{26.5} & 44.7 & 14.2 
		\\
		\hline
		VGG~\cite{bian-arxiv2017} & Kinetics-400 & - & 21.6 & 39.4 & N/A
		\\
		\textbf{Audio-only} (ours)  & Kinetics-400 & - & \textbf{24.8} & \textbf{43.3} & 14.2 
	\end{tabular}
	\vspace{1pt}
	\caption{\textbf{Results of Audio-only models.} VGG* model results are taken from ``iTXN'' submission from Baidu Research to ActivityNet challenge, as documented in this report~\cite{ghanem-activitynet2018}.}
	\label{table:audioonly}
\end{table}

\subsection{Results: Classification \& Detection Analysis}  \label{sec:results_analysis}

\paragraph{Per-class analysis on Kinetics}
Comparing AVSlowFast to SlowFast (77.0\% \vs 75.6\% for 4\x16, R50 backbone), classes that benefited most from audio include [``{\it dancing macarena}'' +24.5\%], [``{\it whistling}'' +24.0\%], [``{\it beatboxing}'' +20.4\%], [``{\it salsa dancing}'' +19.1\%] and [``{\it singing}'' +16.0\%], etc. Clearly, all these classes have distinct sound signatures to be recognized. On the other hand, classes like [``{\it skiing (not slalom or crosscountry)}'' -12.3\%], [``{\it triple jump}'' -12.2\%], [``{\it dodgeball}'' -10.2\%] and [``{\it massaging legs}'' -10.2\%] have the largest performance loss, as sound of these classes tend to be much less correlated the action. 

\paragraph{Per-class analysis on AVA}
We compare per-class results of AVSlowFast to its SlowFast counterparts in Fig.~\ref{fig:ava}. As mentioned in the main paper, classes with largest absolute gain (marked with bold black font) are ``{\it swim}'', ``{\it dance}'', ``{\it shoot}'', ``{\it hit (an object)}'' and ``{\it cut}''. Further, the classes ``{\it push (an object)}'' (3.2\x) and ``{\it throw}'' (2.0\x) largely benefit from audio in relative terms (marked with orange font in Fig.~\ref{fig:ava}). As expected, all these classes have strong sound signature that are easy to recognize from audio. On the other hand, the largest performance loss arises for classes such as ``watch (e.g., TV)'', ``read'', ``eat'' and ``work on a computer'', which either do not have a distinct sound signature (``read'', ``work on a computer'') or have strong background noise sound (``watch (e.g., TV)''). 
We believe explicitly modeling foreground and background sound might be a fruitful future direction to alleviate these challenges. 

\subsection{Details: Kinetics Action Classification}\label{sec:kinetics_details}

We train our models on Kinetics from scratch without any pretraining. Our training and testing closely follows~\cite{slowfast}. We use a synchronous SGD optimizer and train with 128 GPUs using the recipe in \cite{goyal-arxiv2017}. The mini-batch size is 8 clips per GPU (so the total mini-batch size is 1024). The initial base learning rate $\eta$ is 1.6 and we decrease the it according to half-period cosine schedule~\cite{loshchilov-arxiv2016}: the learning rate at the $n$-th iteration is $\eta\cdot0.5[\cos(\frac{n}{n_\text{max}}\pi)+1]$, where $n_\text{max}$ is the maximum training iterations. We adopt a linear warm-up schedule~\cite{goyal-arxiv2017} for the first 8k iterations. 
We use a scale jittering range of [256, 340] pixels for R101 model to improve generalization~\cite{slowfast}. To aid convergence, we initialize all models that use Non-Local blocks (NL) from their counterparts that are trained without NL. We only use NL on res$_4$ (instead of res$_3$+res$_4$ used in~\cite{wang-nonlocal2018}). 

We train with Batch Normalization (BN)~\cite{ioffe-batchnorm2015}, and the BN statistics are computed within each 8 clips.  Dropout~\cite{hinton-dropout2012} with rate 0.5 is used before the final classifier layer.
In total, we train for 256 epochs (60k iterations with batch size 1024, for $\app$240k Kinetics videos) when $T\leq$ 4 frames, and 196 epochs when the Slow pathway has $T>$ 4 frames: it is sufficient to train shorter when a clip has more frames. We use momentum of 0.9 and weight decay of 10$^\text{-4}$.

\subsection{Details: EPIC-Kitchens Classification}\label{sec:epic_details}
We fine-tune from Kinetics pretrained AVSlowFast 8\x8, R101 (w/o NL) for this experiment. For fine-tuning, we freeze all BNs by converting them into affine layers. We train using a single machine with 8 GPUs. Initial base learning rate $\eta$ is set to 0.01 and 0.0006 for verb and noun. We train with batch size 32 for 24k and 30k for verb and noun respectively. We use a step wise decay of the learning rate by a factor of 10\x~at 2/3 and 5/6 of full training.  For simplicity, we only use a single center crop for testing. 

\subsection{Details: Charades Action Classification}\label{sec:charades_details}
We fine-tune from the Kinetics pretrained AVSlowFast 16\x8, R101 + NL model, to account for the longer activity range of this dataset, and a per-class sigmoid output is used to account for the mutli-class nature of the data. We train on a single machine (8 GPUs) for 40k iterations using a batch size of 8 and a base learning rate $\eta$  of 0.07 with one 10\x~decay after 32k iterations. We use a Dropout rate of 0.7. For inference, we temporally max-pool scores \cite{wang-nonlocal2018,slowfast}. All other settings are the same as those of Kinetics. 

\subsection{Details: AVA Action Detection}\label{sec:ava_details}

We follow the detection architecture introduced in~\cite{slowfast}, which is adapted from Faster R-CNN~\cite{ren-nips2015} for video. 
Specifically, we set the spatial stride of res$_5$ from 2 to 1, thus increasing the spatial resolution of res$_5$ by 2\x. RoI features are then computed by applying RoIAlign~\cite{he-maskrcnn2017} spatially and global average pooling temporally. These features are then fed to a per-class, sigmoid-based classifier for multi-label prediction. Again, we initialize from Kinetics pretrained models and train 52k iterations with initial learning rate $\eta$ of 0.4 and batch size 16 (we train across 16 machines, so effective batch size 16\x16=256).  We pre-compute proposals using an off-the-shelf Faster R-CNN person detector with ResNeXt-101-FPN backbone. It is pretrained on ImageNet and the COCO human keypoint data and more details can be found in \cite{slowfast,wu-lfb2019}. 

\subsection{Details: Self-supervised Evaluation}\label{sec:ssl_details}
For self-supervised pretraining, we train on Kinetics-400 for 120k iterations with per-machine batch size 64 across 16 machines and initial learning rate 1.6, similar to \S\ref{sec:kinetics_details}, but with step-wise schedule. The learning rate is decayed with 10\x~three times at 80k, 100k and 110k iterations. We use linear warm-up (starting from learning rate 0.001) for the first 10k iterations. As noted in \sref{sec:ssl}, we adopt the curriculum learning idea for audiovisual synchronization~\cite{korbar-nips2018} to first train with easy negatives for the first 60k iterations and then switch to a mix of easy and hard negatives (1$/$4 hard, 3$/$4 easy) for the remaining 60k iterations. The easy negatives com from different videos, while hard negatives have a temporal displacement of at least 0.5 seconds. 

For the ``linear classification protocol'' experiments on UCF and HMDB, we train 320k iterations (echoing~\cite{kolesnikov-ssl2019}, we found it beneficial to train long iterations in this setting) with an initial learning rate of 0.01, a half-period cosine decay schedule  and a batch size of 64 on a single machine with 8~GPUs. For the ``train all layers'' setting, we train 80k $/$ 30k iterations with batch size 16 (also on a single machine), an initial learning rate of 0.005 $/$ 0.01 and a half-period cosine decay schedule, for UCF and HMDB, respectively. 

\section{Details: Kinetics-Sound dataset} \label{sec:ks_details}

The original 34 classes selected in~\cite{arandjelovic-iccv2017} are based on an earlier version of the Kinetics dataset. Some classes are removed since then. Therefore, we use the following 32 classes that are kept in current version of Kinetics-400 dataset: ``blowing nose'', ``blowing out candles'', ``bowling'', ``chopping wood'', ``dribbling basketball'', ``laughing'', ``mowing lawn'', ``playing accordion'', ``playing bagpipes'', ``playing bass guitar'', ``playing clarinet'', ``playing drums'', ``playing guitar'', ``playing harmonica'', ``playing keyboard'', ``playing organ'', ``playing piano'', ``playing saxophone'', ``playing trombone'', ``playing trumpet'', ``playing violin'', ``playing xylophone'', ``ripping paper'', ``shoveling snow'', ``shuffling cards'', ``singing'', ``stomping grapes'', ``strumming guitar'', ``tap dancing'', ``tapping guitar'', ``tapping pen'', ``tickling''.

{\small
\bibliographystyle{ieee_fullname}
\bibliography{strings,bibs}
}

\end{document}